\documentclass{article}
\usepackage[square,numbers]{natbib}
\usepackage{graphicx} 
\usepackage[table]{xcolor}
\graphicspath{ {images/} }
\usepackage{subcaption}
\usepackage{amsmath}

\usepackage[colorinlistoftodos]{todonotes}
\usepackage[colorlinks=true, allcolors=blue]{hyperref}

\usepackage[final]{nips_2016}

\usepackage[utf8]{inputenc} 
\usepackage[T1]{fontenc}    
\usepackage{hyperref}       
\usepackage{url}            

\usepackage{booktabs}       
\usepackage{amsfonts}       
\usepackage{nicefrac}       
\usepackage{microtype}      
\newcommand{\minisection}[1]{\vspace{0.04in} \noindent {\bf #1}\ \ }

\title{Ensembles of Generative Adversarial Networks}

\author{
  Yaxing Wang, Lichao Zhang, Joost van de Weijer\\
  Computer Vision Center\\
  Barcelona, Spain \\
  \texttt{\{yaxing,lichao,joost\}@cvc.uab.es}}

\begin{document}

\maketitle

\begin{abstract}
Ensembles are a popular way to improve results of discriminative CNNs. The combination of several networks trained starting from different initializations improves results significantly. In this paper we investigate the usage of ensembles of GANs. The specific nature of GANs opens up several new ways to construct ensembles. The first one is based on the fact that in the minimax game which is played to optimize the GAN objective the generator network keeps on changing even after the network can be considered optimal. As such ensembles of GANs can be constructed based on the same network initialization but just taking models which have different amount of iterations. These so-called self ensembles are much faster to train than traditional ensembles. The second method, called cascade GANs, redirects part of the training data which is badly modeled by the first GAN to another GAN. In experiments on the CIFAR10 dataset we show that ensembles of GANs obtain model probability distributions which better model the data distribution. In addition, we show that these improved results can be obtained at little additional computational cost. 
\end{abstract}

\section{Introduction}



Unsupervised learning extracts features from the unlabeled data to describe hidden structure, which is arguably more attractive, compelling and challenging than supervised learning. One unsupervised application which has gained momentum in recent years, is the task to generate images. The most common image generation models fall into two main approaches. The first one is based on probabilistic generative models, which includes autoencoders\cite{rumelhart1985learning} and powerful variants\cite{vincent2008extracting,bengio2007greedy,vincent2010stacked}. The second class, which is the focus of this paper, is called Generative Adversarial Networks (GANs)\cite{goodfellow2014generative}. These networks combine a generative network and a discriminative network. The advantage of these networks is that they can be trained with back propagation. In addition, since the discriminator network is a convolutional network, these networks are optimizing an objective which reflects human perception of images (something which is not true when minimizing a Euclidean reconstruction error).


Since their introduction GANs have been applied to a wide range of applications and several improvements have been proposed. Ranford et al.~\cite{radford2015unsupervised} propose and evaluate several constraints on the network architecture, thereby improving significantly the stability during training. They call this class of GANs, Deep Convolutional GANs (DCGAN), and we will use these GANs in our experiments. Denton et al.~\cite{denton2015deep} propose a Laplacian pyramid framework based on a cascade of convolutional networks to synthesize images at multiple resolutions. Further improvements on stability and sythesized quality have been proposed in~\cite{chen2016infogan,donahue2016adversarial,im2016generating,salimans2016improved}. 



Several works have shown that, for disciminatively trained CNNs, applying an ensemble of networks is a straightforward way to improve results~\cite{krizhevsky2012imagenet,wang2015unsupervised,zeiler2014visualizing}. The ensemble is formed by training several instances of a network from different initializations on the same dataset, and combining them e.g. by a simple probability averaging. Krizhevsky et al.~\cite{krizhevsky2012imagenet} applied seven networks to improve results for image classification on ImageNet. Similarly, Wang and Gupta~\cite{wang2015unsupervised} showed a significant increase in performance using an ensembles of three CNNs for object detection. These works show that ensembles are a relatively easy (be it computationally expensive) way to improve results. To the best of our knowledge the usage of ensembles has not yet been evaluated for GANs. Here we investigate if ensembles of GANs generate model distributions which closer resembles the data distribution.

We investigate several strategies to train an ensemble of GANs. Similar as~\cite{krizhevsky2012imagenet} we train several GANs from scratch from the data and combine them to generate the data (we will refer to these as standard ensembles). When training GANs the minimax game prevents the networks from converging, but instead the networks $G$ and $D$ constantly remain changing. Comparing images generated by successive epochs shows that even after many epochs these networks generate significantly different images. Based on this observation we propose self-ensembles which are generated by several models $G$ which only differ in the number of training iterations but originate from the same network initialization. This has the advantage that they can be trained much faster than standard ensembles. 

In a recent study on the difficulty of evaluating generative models, Theis et al.~\cite{theis2015note} pointed out the danger that GANs could be quite accurately modeling part of the data distribution while completely failing to model other parts of the data. This problem would not easily show up by an inspection of the visual quality of the generated examples. The fact that the score of the discriminative network $D$ for these not-modelled regions is expected to be high (images in these regions would be easy to recognize as coming from the true data because there are no similar images generated by the generative network) is the bases of the third ensemble method we evaluate. This method we call a cascade ensemble of GANs. We redirect part of the data which is badly modelled by the generative network $G$ to a second GAN which can then concentrate on generating images according to this distribution.  We evaluate results of ensemble of GANs on the CIFAR10 dataset, and show that when evaluated for image retrieval, ensembles of GANs have a lower average distance to query images from the test set, indicating that they better model the data distribution.  



\section{Generative Adversarial Network}
A GAN is a framework consisting of a deep generative model $G$ and a discriminative model $D$, both of which play a minimax game. The aim of the generator is to generate a distribution $p_g$ that is similar to the real data distribution $p_{data}$ such that the discriminative network cannot distinguish between the images from the real distribution and the ones which are generated (the model distribution).

\begin{figure}[tb]
\begin{subfigure}{0.5\textwidth}
\includegraphics[width=\linewidth]{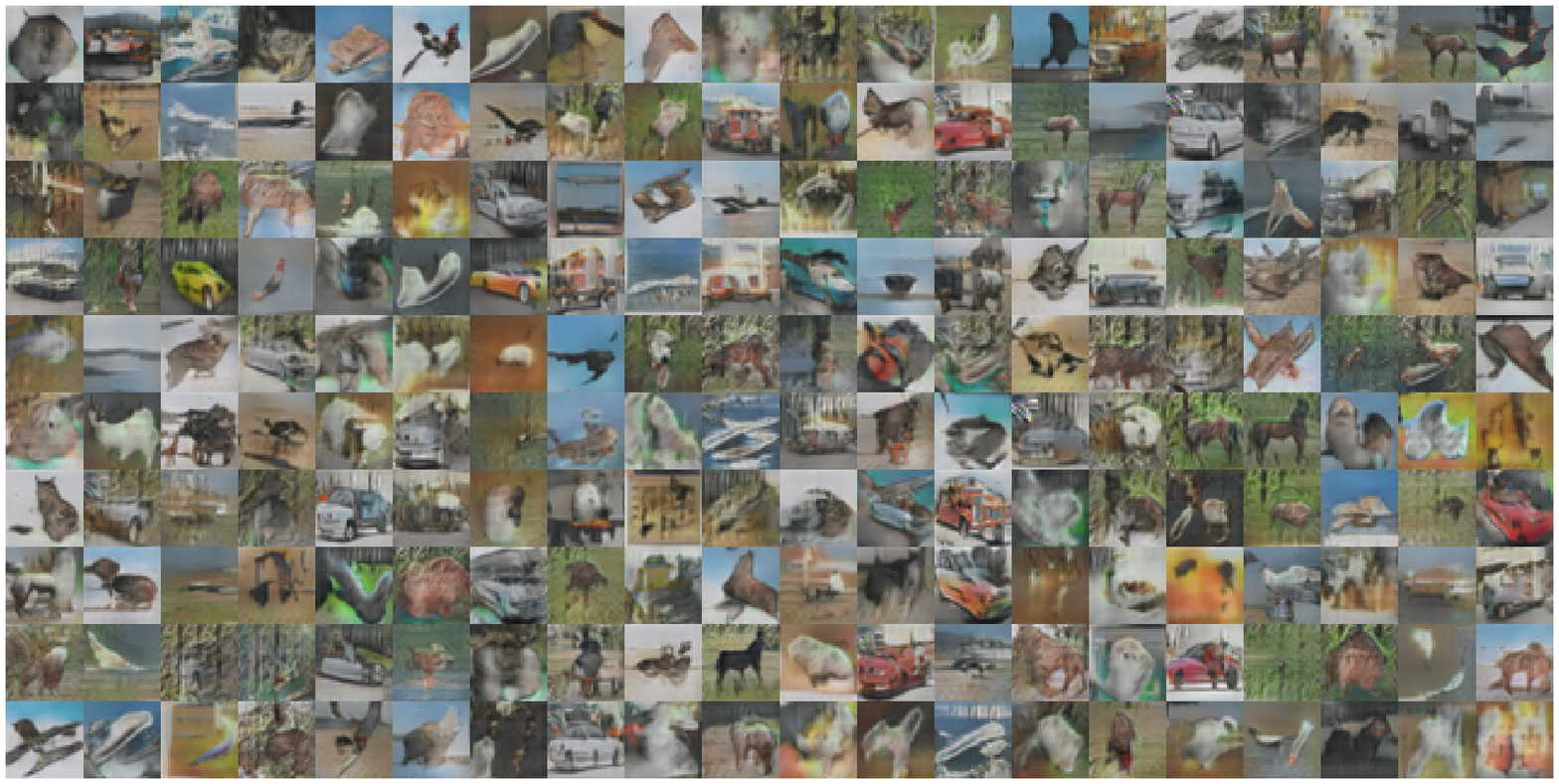}
\end{subfigure}
\hspace*{\fill} 
\begin{subfigure}{0.5\textwidth}
\includegraphics[width=\linewidth]{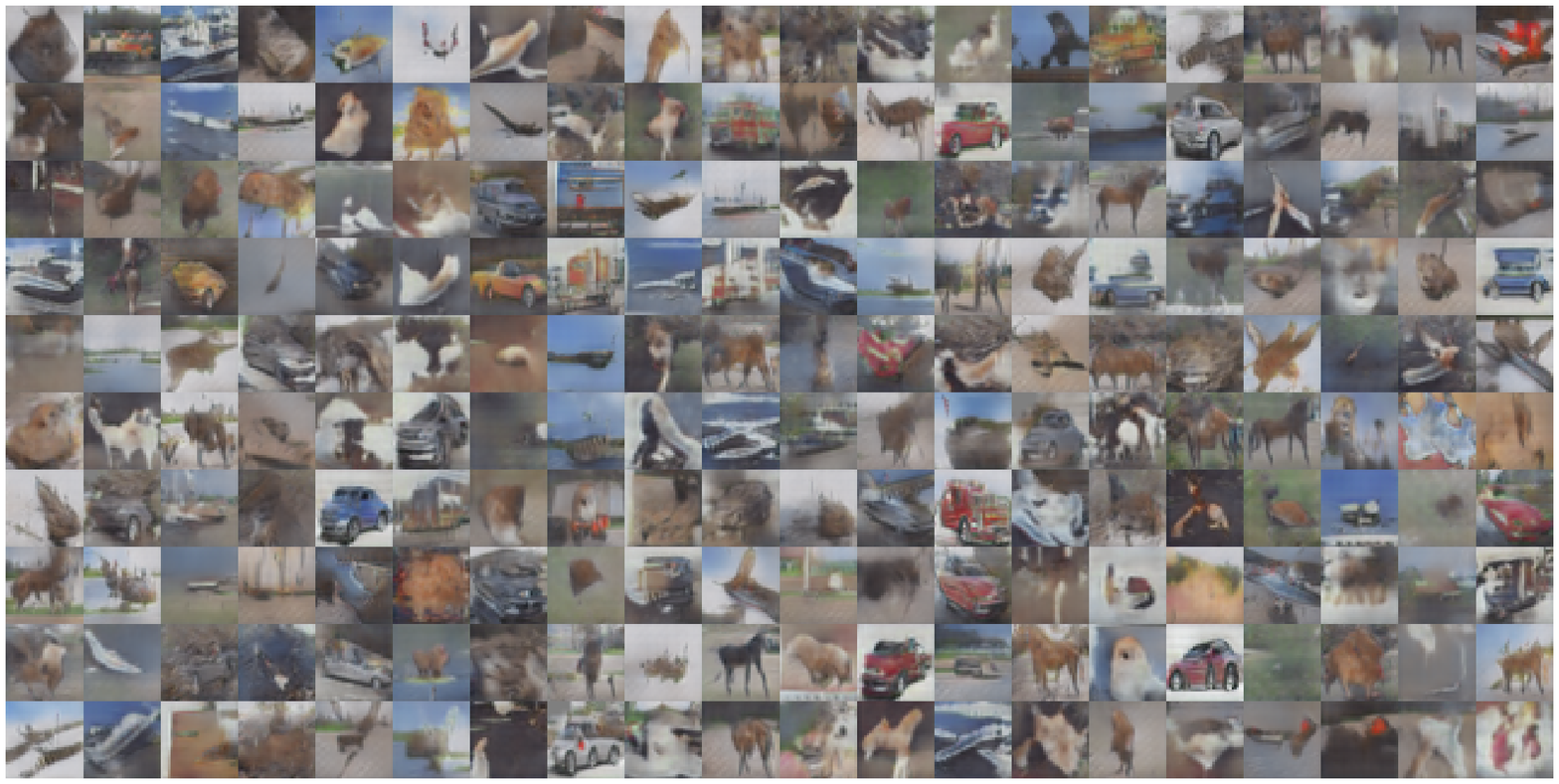}
\end{subfigure}
\caption{Two-hundred images generated from the same random noise with DCGAN on CIFAR dataset after 72 (left) and 73 (right) epochs of training from same network initialization. In the minimax game the generator and discriminator keep on changing. The resulting distributions $p_g$ are clearly different (for example very few saturated greens in the images on the right). } \label{fig:GANepochs}
\end{figure}

Let $x$ be a real image drawn from the real data distribution $p_{data}$ and $z$ be random noise. The noise variable $z$ is transformed into a sample $G(z)$ by a generator network $G$  which synthesizes samples from the distribution $p_g$. The discriminative model $D(x)$ computes the probability that input data $x$ is from $p_{data}$ rather than from the generated model distribution $p_g$.  Ideally $D(x) = 0 $ if $x\sim p_g$ and $D(x) = 1$ if $x\sim p_{data}$. More formally, the generative model and discriminative model are trained by solving:
\begin{equation}
\underset{G}{\text{min}}\ \underset{D}{\text{max}}\ V\left ( D, G\right) = E_{x\sim p_{data}}\left [ \log D\left ( x \right ) \right ] + E_{z\sim noise}\left [ \log\left( 1-D\left( G(z) \right) \right) \right ]
\end{equation}
In our implementation of GAN we will use the DCGAN\cite{radford2015unsupervised} which improved the quality of GANs by the usage of strided convolutions  and fractional-strided convolutions instead of pooling layers in  both  generator and discriminator, as well as the RLlu and leakyReLu activation function.

We shortly describe two observations which are particular to GANs and which are the motivations for the ensemble models we discuss in the next section. \\

\minisection{Observation 1:} In Fig \ref{fig:GANepochs} we show the images which are generated by a DCGAN for two successive epochs. It can be observed that the generated images change significantly in overall appearance from one epoch to the other (from visual inspection quality of images does not increase after epoch 30 but the overall appearance still varies considerably). The change is caused by the fact that in the minimax game the generator and the discriminator constantly vary and do not converge in the sense that discriminatively trained networks do. Rather than a single generator and discriminator one could consider the GAN training process to generate a set of generative-discriminative network pairs. Given this observation it seems sub-optimal to choose a single generator network from this set to generate images from, and ways to combine them should be explored.

\minisection{Observation 2:} A drawback of GANs as pointed out be Theis et al.~\cite{theis2015note} is that they potentially do not describe the whole data distribution $p_{data}$. The reasoning is based on the observation that objective function of GANs have some resemblance with the Jensen-Shannon divergence (JSD). They show that for a very simple bi-modal distribution, minimizing the JSD yields a good fit to the principal mode but ignores other parts of the data. For the application of generating images with GANs this would mean that for part of the data distribution the model does not generate any resembling images.

\begin{figure}[tb]
\centering
\includegraphics[width=.6\textwidth]{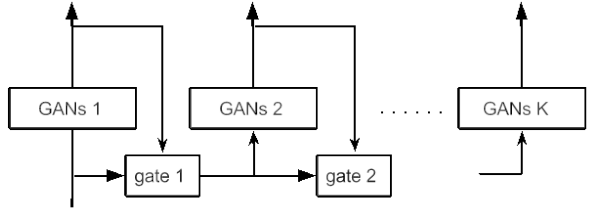}
\caption{\label{fig:cGANs} The proposed cGANs framework consists of multiple GANs. We start with all train data (left side) and train the first GAN until no further improvements are obtained. We then use the gate-function to select part of the train data to be modeled by the second GAN, etc}
\end{figure}

\section{Ensembles of Generative Adversarial Networks}
As explained in the introduction we investigate the usage of ensembles of GANs and evaluate their performance gain. Based on the observations above we propose three different approaches to construct an ensemble of GANs. The aim of the proposed schemes is to obtain a better estimation of the real data distribution $p_{data}$. 

\minisection{Standard Ensemble of GANs (eGANs):}
We first consider a straightforward extension of the usage of ensembles to GANs. This is similar to ensembles used for discriminative CNNs which have shown to result in significant performance gains~\cite{krizhevsky2012imagenet,wang2015unsupervised,zeiler2014visualizing}. Instead of training a single GAN model on the data, one trains a set of GAN models from scratch from a random initialization of the parameters. When generating data one randomly chooses one of the GAN models and then generates the data according to that model.

\minisection{Self-ensemble of GANs (seGANs):}Other than discriminative networks which minimize an objective function, in a GAN the min/max game results in a continuing shifting of the generative and discriminative network (see also observation 1 above). An seGAN exploits this fact by combining models which are based on the same initialization of the parameters but only differ in the number of training iterations. This would have the advantage over eGANs that it is not necessary to train each GAN in the ensemble from scratch. As a consequence it is much faster to train seGANs than eGANs.

\minisection{Cascade of GANs (cGANs):} The cGANs is designed to address the problem described in observation 2; part of the data distribution might be ignored by the GAN. The cGANs framework as illustrated in Figure \ref{fig:cGANs} is designed to train GANs to effectively push the generator to capture the whole distribution of the data instead of focusing on the main mode of the density distribution.  It consists of multiple GANs and gates. Each of the GAN trains a generator to capture the current input data distribution which was badly modeled by  previous GANs. To select the data which is re-directed to the next GAN we use the fact that for badly modeled data $x$, the discriminator value $D(x)$ is expected to be high. When $D(x)$ is high this means that the discriminator is confident this is real data, which most probably is caused by the fact that there are few generated examples $G(z)$ nearby. We use the gate-function $Q$ to re-direct the data to the next GAN according to:
\begin{equation}
Q(x_k)=\left\{\begin{matrix}
1   & \text{if}\ D(x) > t_r\\ 
 0& \text{else}
\end{matrix}\right.\label{Eq:gate}
\end{equation}
where $Q(x_k)=1$ means that $x$ will be used to train the next adversarial network. In practice we will train a GAN until satisfactory results are obtained. Then evaluate Eq.~\ref{Eq:gate} and train the following GAN with the selected data, etc. We set a ratio $r$ of images which are re-directed to the next GAN, and select the threshold $t_r$ accordingly. In the experiments we show results for several ratio settings. 



\begin{figure}[tb]
\centering
\captionsetup{justification=centering}
\includegraphics[width=0.4\textwidth]{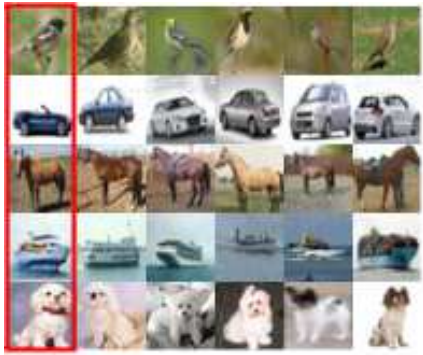}
\caption{\label{fig:retrieval} Retrieval example. The leftmost column, annotated by a red rectangle, includes five query images from the test set. To the right the five nearest neighbors in the training set are given.}
\end{figure}

\section{Experiments}

\subsection{Experimental setup}
The evaluation of generative methods is known to be problematic\cite{theis2015note}. Since we are evaluating GANs which are based on a similar network architecture (we use the standard settings of DCGAN\cite{radford2015unsupervised}), the quality of the generated images is similar and therefore uninformative as an evaluation measure. Instead, we are especially interested to measure if the ensembles of GANs better model the data distribution. 

To measure this we propose an image retrieval experiment. We represent all images, both from a held-out test set as well as generated images by the GANs, with an image descriptor based on discriminatively trained CNNs. For all images in the test dataset we look at their nearest neighbor in the generated image dataset. Comparing ensemble methods based on these nearest neighbor distances allows us to assess the quality of these methods. We are especially interested if some images in the dataset are badly modeled by the network, which would lead to high nearest neighbor distances. At the end of this section we discuss several evaluation criteria based on these nearest neighbor distances. 

For the representation of the images we finetune an Alexnet model (pre-trained on ImageNet) on the CIFAR10 dataset. It has been shown that the layers from AlexNet describe images at varying level of semantic abstraction  \cite{zeiler2014visualizing}; the lower  layers  of  the  neural network  mainly capture low-level information, such as colors, edges and corners etc, whereas the upper layers contain more semantic features like heads, wheels, etc. Therefore, we combine the output of the first convolutional layer, the first fully connected layer and the final results after the softmax layer into one image representation. The conv1 layer is grouped into a 3x3 spatial grid, resulting in a $3\times3\times96=864$ dimensional vector. For the nearest neighbor we use the Euclidean distance\footnote{The average distance between images in the dataset is normalized to be one for each of the three parts conv1, fc7, and prob}. Example retrieval results with this system are provided in Fig.~\ref{fig:retrieval}, where we show the five nearest neighbors for several images. It shows that the image representation captures both color and texture of the image, as well as semantic content. In the experiments we will use the retrieval system to compare various ensembles of GANs. 



\minisection{Evaluation criteria:} To evaluate the quality of the retrieval results, we will consider two measures. As mentioned they are based on evaluating the nearest neighbor distances of the generated images to images in the CIFAR testset. Consider $d_{i,j}^k$  to be 
the distance of the $j^{th}$ nearest image generated by method $k$ to test (query) image $i$, and ${\bf{d}}^k_{j}  = \left\{ {d_{1,j}^k ...d_{n,j}^k } \right\}$ the set of $j^{th}$-nearest distances to all $n$ test images. Then the Wilcoxon signed-rank test (which we will only apply for the nearest neighbor $j=1$),  is used to test the hypothesis that the median of the difference between two nearest distance distributions of generators is zero, in which case they are equally good (i.e., the median of the distribution ${\bf{d}}^k_{1}-{\bf{d}}^m_{1}$ when considering generator $k$ and $m$). If they are not equal the test can be used to assess which method is statistically better. This method is for example popular to compare illuminant estimation methods~\cite{hordley2006reevaluation}.


For the second evaluation criterion, consider ${\bf{d}}^t_{j}$ to be the distribution of the $j^{th}$ nearest distance of the train images to the test dataset. Since we consider that the train and test set are drawn from the same dataset, the distribution ${\bf{d}}^t_{j}$ can be considered the optimal distribution which a generator could attain (considering it generates an equal amount of images as present in the trainset). To model the difference with this ideal distribution we will consider the relative increase in mean nearest neighbor distance given by:
\begin{equation}
\hat{d}_{j}^{k} = \frac{ \bar{d}_{j}^{k}- \bar{d}_{j}^{t}}{ \bar{d}_{j}^{t}}
\end{equation}
where
\begin{equation}
\bar{d}_{j}^{k} = \frac{1}{N}\sum_{i = 1}^{N}d_{i,j}^{k},\;\;\;\;\;\;\;\;\;\;\;\;\bar{d}_{j}^{t} = \frac{1}{N}\sum_{i = 1}^{N}d_{i,j}^{t}
\end{equation}
and where $N$ is the size of the test dataset. E.g., $\hat{d}_{1}^{GAN}=0.1$ means that for method GAN the average distance to the nearest neighbor of a query image is 10 \% higher than for data drawn from the ideal distribution.

\begin{table}[tb]
\centering
\captionsetup{justification=centering}
\setlength{\arrayrulewidth}{1.8\arrayrulewidth}
\begin{tabular}{|c|c|c|c|c|c|c|c|}
\hline
              & \rotatebox{90}{0. pdata}  &\rotatebox{90}{1. GAN}   &\rotatebox{90}{2. cGANs(0.5)\;}   &\rotatebox{90}{3. cGANs(0.6)\;}   &\rotatebox{90}{4. cGANs(0.7)}  &\rotatebox{90}{5. cGANs(0.8)}  &\rotatebox{90}{6. cGANs(0.9)\;}  \\ \hline
0.           &  \cellcolor{yellow!50}{0}    &\cellcolor{green!50}{1}   &\cellcolor{green!50}{1}  & \cellcolor{green!50}{1}   &\cellcolor{green!50}{1}    &\cellcolor{green!50}{1}     &\cellcolor{green!50}{1}     \\ \hline
1.                & \cellcolor{red!50}{-1}    &\cellcolor{yellow!50}{0}   & \cellcolor{red!50}{-1}  & \cellcolor{red!50}{-1}  &\cellcolor{red!50}{-1}   &\cellcolor{red!50}{-1}    &\cellcolor{red!50}{-1}     \\ \hline
2.                &  \cellcolor{red!50}{-1}    &\cellcolor{green!50}{1}  &\cellcolor{yellow!50}{0}  & \cellcolor{red!50}{-1}  &\cellcolor{red!50}{-1}   &\cellcolor{red!50}{-1}    &\cellcolor{red!50}{-1}     \\ \hline
3.       & \cellcolor{red!50}{-1}    &\cellcolor{green!50}{1}   &\cellcolor{green!50}{1}  & \cellcolor{yellow!50}{0}   &\cellcolor{red!50}{-1}     &\cellcolor{red!50}{-1}    &\cellcolor{red!50}{-1}      \\ \hline
4.               & \cellcolor{red!50}{-1}    &\cellcolor{green!50}{1}   &\cellcolor{green!50}{1}  &\cellcolor{green!50}{1}  &\cellcolor{yellow!50}{0}    &\cellcolor{yellow!50}{0}    &\cellcolor{green!50}{1}     \\ \hline
5.          & \cellcolor{red!50}{-1}    &\cellcolor{green!50}{1}   &\cellcolor{green!50}{1}  &\cellcolor{green!50}{1}   &\cellcolor{yellow!50}{0}     &\cellcolor{yellow!50}{0}     &\cellcolor{green!50}{1}     \\ \hline
6.              & \cellcolor{red!50}{-1}   &\cellcolor{green!50}{1}    &\cellcolor{green!50}{1}  &\cellcolor{green!50}{1}  &\cellcolor{red!50}{-1}   &\cellcolor{red!50}{-1} &\cellcolor{yellow!50}{0}     \\ \hline
\end{tabular}
\caption{Wilcoxon signed-rank test for cGANs approach. The number between brackets refers to the ratio $r$ which is varied. Best results are obtained with $r$ equal to 0.7 and 0.8.}
\label{Wilcoxon_cGANs}
\end{table}

\begin{table}[tb]
\centering
\captionsetup{justification=centering}
\setlength{\arrayrulewidth}{1.8\arrayrulewidth}
\begin{subtable}{.5\textwidth}
\centering
\begin{tabular}{|c|c|c|c|c|}
\hline
  &\rotatebox{90}{1. cGANs\;}   &\rotatebox{90}{2. eGANs}   &\rotatebox{90}{3. seGANs}  \\ \hline

1.        &\cellcolor{yellow!50}{0/10/0}&\cellcolor{red!50}{1/0/9}  & \cellcolor{red!50}{0/1/9}      \\ \hline
2.    &\cellcolor{green!50}{9/0/1}  &\cellcolor{yellow!50}{0/10/0} & \cellcolor{yellow!50}{4/1/5}   \\ \hline
3.       & \cellcolor{green!50}{9/1/0}   &\cellcolor{yellow!50}{5/1/4}  &\cellcolor{yellow!50}{0/10/0}     \\ \hline
\end{tabular}
\caption{}
\end{subtable}%
\begin{subtable}{.5\textwidth}
\centering
\begin{tabular}{|c|c|c|c|c|}
\hline
& \rotatebox{90}{1. GAN\;}   &\rotatebox{90}{2. seGANs(2)} &\rotatebox{90}{3. seGANs(4)}&\rotatebox{90}{4. seGANs(8)\;}  \\ \hline

1.  &\cellcolor{yellow!50}{0/10/0}  &\cellcolor{red!50}{0/0/10}  & \cellcolor{red!50}{0/0/10}  &\cellcolor{red!50}{ 0/0/10}    \\ \hline
2.   &\cellcolor{green!50}{10/0/0} &\cellcolor{yellow!50}{0/10/0} & \cellcolor{red!50}{0/0/10}& \cellcolor{red!50}{0/0/10 } \\ \hline
3.   & \cellcolor{green!50}{10/0/0}   &\cellcolor{green!50}{10/0/0}  &\cellcolor{yellow!50}{ 0/10/0} &\cellcolor{red!50}{0/2/8}    \\ \hline
4.   & \cellcolor{green!50}{10/0/0}   &\cellcolor{green!50}{10/0/0}  &\cellcolor{green!50}{8/2/0} &\cellcolor{yellow!50}{0/10/0}    \\ \hline
\end{tabular}
\caption{}
\end{subtable}%
\caption{Wilcoxon signed-rank test evaluation. Results are shown as (A)/(B)/(C) where A, B and C are appearing times of 1, 0 and -1 respectively during 10 experiments. The more 1 appears, the better the method. In between brackets we show the number of GAN networks in the ensemble (a) shows eGANs and seGANs outperform cGANs; and the more models used in the ensemble the better the seGANs is shown in (b).}
\label{Wilcoxon_different_GANs}
\end{table}

\subsection{Results}
To evaluate the different configuration for ensembles of GANs we perform several experiments on the CIFAR10 dataset. This dataset has 10 different classes, 50000 train images and 10000 test images of size $32\times32$. In our experiments we compare various generative models. With each of them we generate 10000 images and perform the evaluations discussed in the previous section. 

A cGANs has one parameter, namely the ratio $r$ of images which will be diverted to the second GAN, which we evaluate in the first experiment. The results of the signed-rank test for several different settings of $r$ are provided in Table~\ref{Wilcoxon_cGANs}. In this table, a zero refers to no statistical difference between the distributions. A one (or minus one) refer to non-zero median of the difference of the distributions, indicating that the method is better (or worse) than the method to which it is compared. 

In the graph we have also included the training dataset and a single GAN. For a fair comparison we only consider 10.000 randomly selected images from the training dataset (similar to the number of images which are generated by the generative models). As expected, the distribution of minimal distances to the test images of the training dataset is superior to any of the generative models.  We see this as an indication that the retrieval system is a valid method to evaluate generative models. Next, in Table~\ref{Wilcoxon_cGANs}, we can see that, independent of $r$, cGANs always obtains superior results to using a single standard GAN. Finally, the results show that the best results are obtained when diverting images to the second GAN with a ratio of 0.7 or 0.8. In the rest of the experiments we fix $r=0.8$.

In the next experiment we compare the different approaches to ensembles of GANs. We start by only combining two GANs into each of the ensembles. We have repeated the experiments 10 times and show the results in Table~\ref{Wilcoxon_different_GANs}a. We found that the results of GAN did not further improve after 30 epochs of training. We therefore use 30 epochs for all our trained models. For seGANs we randomly pick models between 30 and 40 epochs of training. The cGANs obtain significantly less results than eGANs and seGANs. Interestingly, the seGANs obtains similar results as eGANs. Whereas eGANs is obtained by re-training a GAN model from scratch, the seGANs is formed by models starting from the same network initialization and therefore much faster to compute than eGANs.

\begin{figure}[tb]
\begin{subfigure}{0.5\textwidth}
\includegraphics[width=\linewidth]{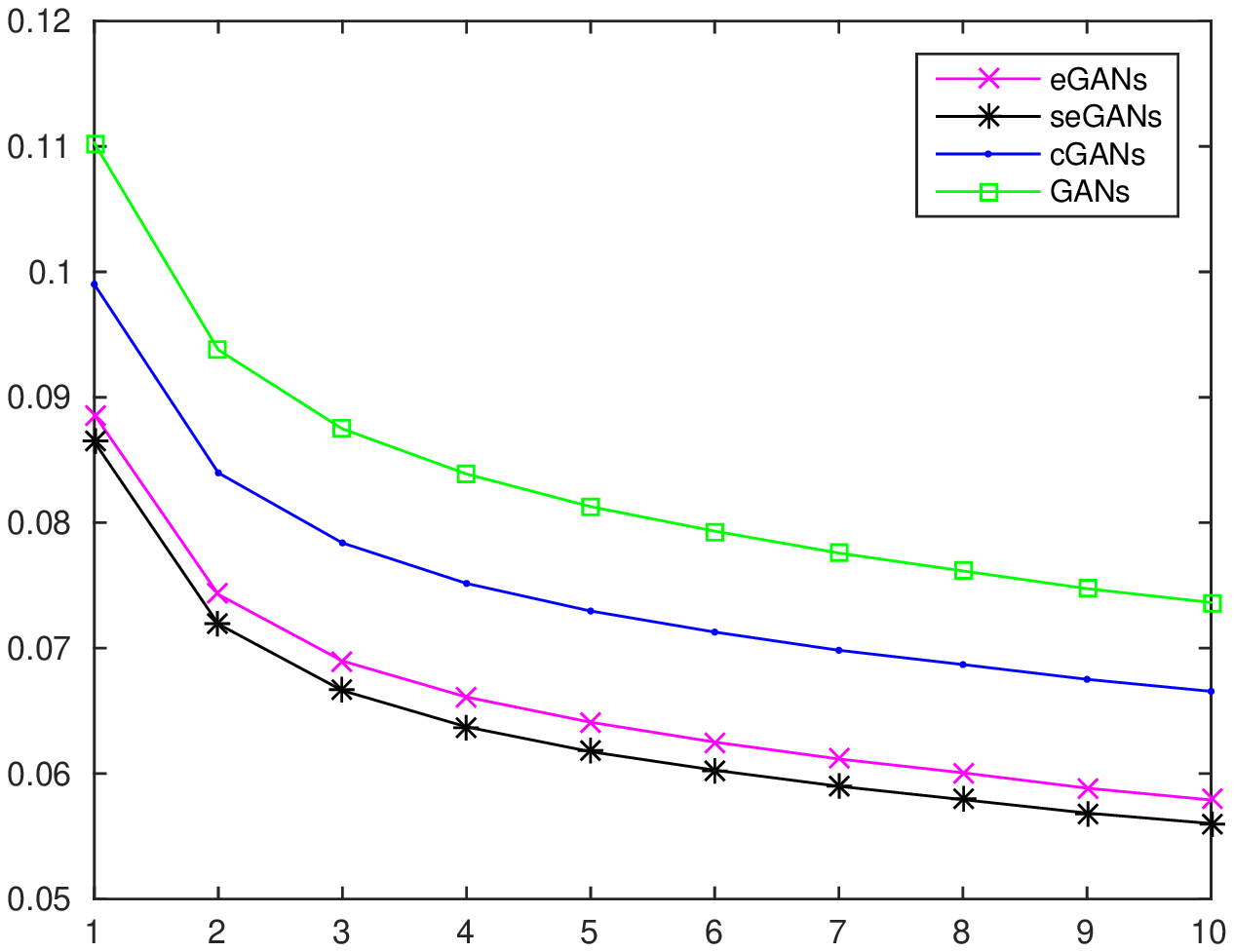}
\end{subfigure}
\hspace*{\fill} 
\begin{subfigure}{0.5\textwidth}
\includegraphics[width=\linewidth]{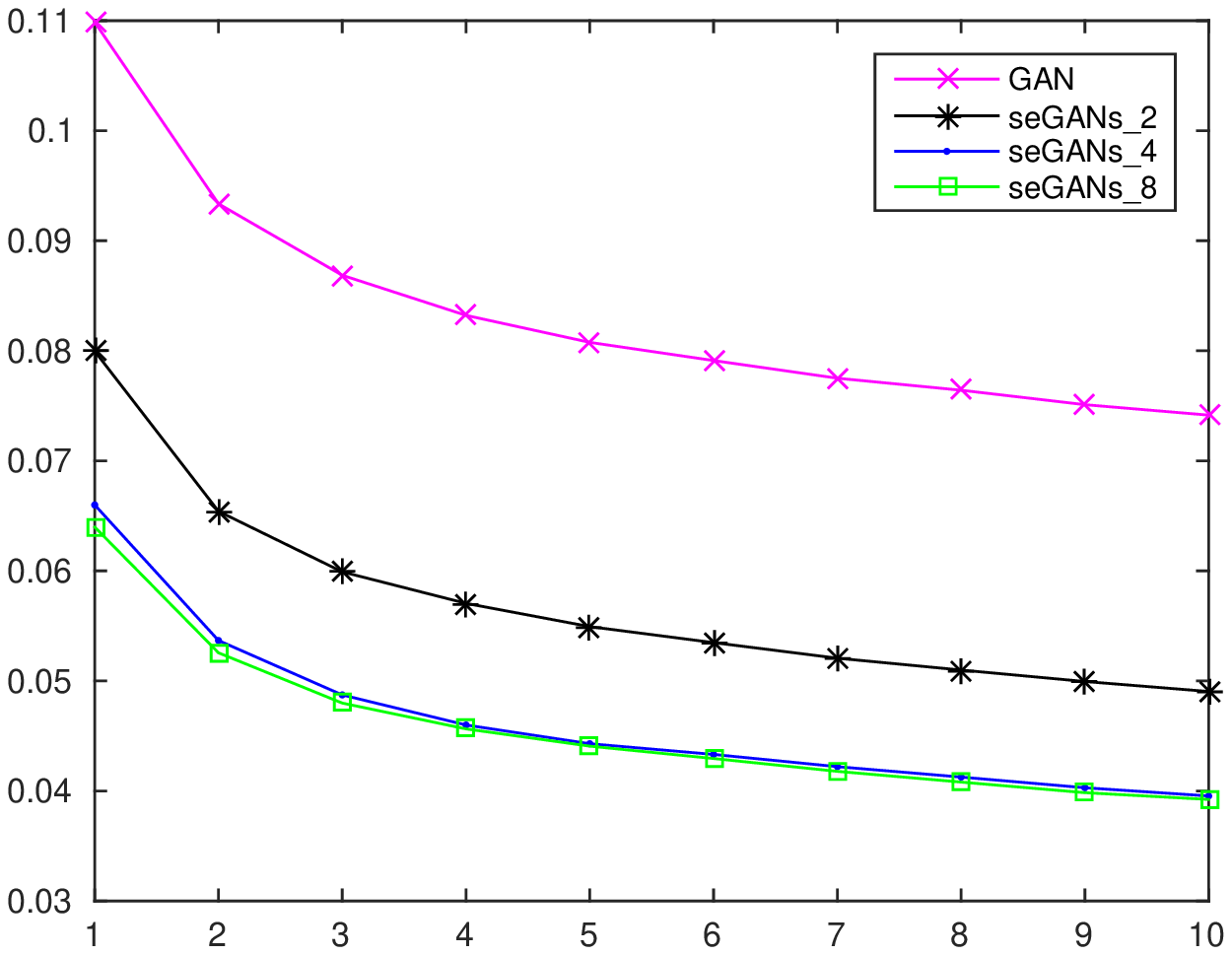}
\end{subfigure}
\caption{(left) Showing $\hat{d}_{j}^{k}$ for $k \in \left\{ {cGANs,eGANs,seGANs}\right\}$ and $j \in \left\{ {1, \ldots ,10} \right\}$ along the horizontal axes. (right) similar but for seGANs with varying number of models.} \label{fig:GANepochs2}
\end{figure}

The results for the ensembles of GANs are also evaluated with the average increase in nearest neighbor distance in Fig~\ref{fig:GANepochs2}(left). In this plot we consider not only the closest nearest neighbor distance, but also the k-nearest neighbors (horizontal axis).  All ensemble methods improve over using just a single GAN. This evaluation measure shows that seGANs obtains similar results as eGANs again. 

\begin{figure}[tb]
\centering
\captionsetup{justification=centering}
\includegraphics[width=1\textwidth]{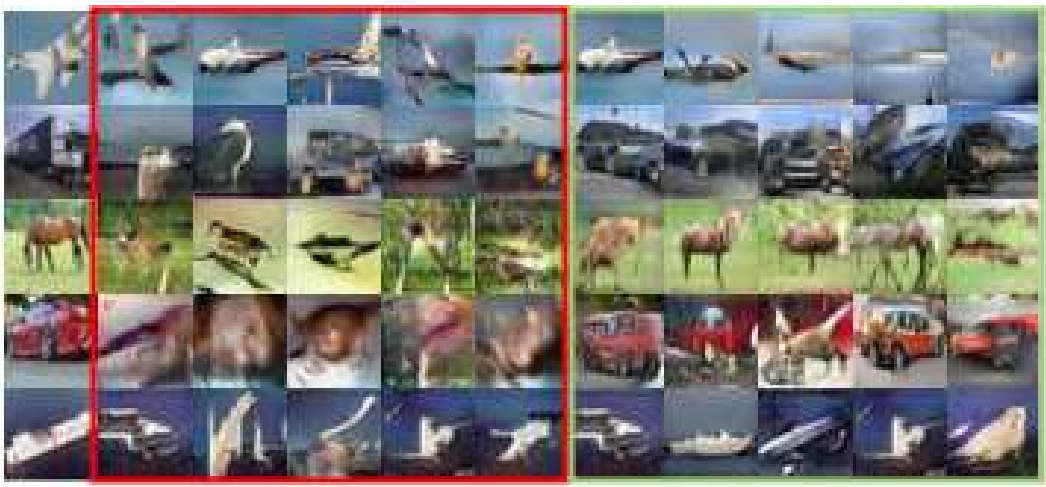}
\caption{Visualization of samples from single GAN and seGANs(4). The leftmost column is the query image from the test dataset; The red box shows the 5 nearest examples from GAN; and the green box shows examples from seGANs(4)\textbf{}
the neighboring sample}
\label{fig:generated_image}
\end{figure}

Finally, we have run seGANs by combining more than two GANs, we have considered ensembles of 2, 4, 6, and 8 GANs (see  Table~\ref{Wilcoxon_different_GANs}b). And then results are repeated 10 times. Results shown in Fig~\ref{fig:GANepochs2}(right) show that the average increase in k-nearest distance decreases, when increasing the number of networks in the ensemble, but levels off from 4 to 8. We stress that when combine 2, 4 or 8 GANs, we keep the number of generated images to 10.000 like in all experiments. So in the case of seGANs(8) 1250 images are generated from each GAN. The average increase in nearest distance drops from 0.11\% for a single GAN to 0.06\% for  a seGANs combining 8 GANs which is a drop of 40\%. In Fig~\ref{fig:generated_image} we show several examples where a single GAN does not generate any similar image whereas the ensemble GAN does. 




\section{Conclusion}
We have evaluated several approaches to construct ensembles of GANs. The most promising results were obtained by seGANs, which were shown to obtain similar results
as eGANs, but have the advantage that they can be trained with little additional computational cost. Experiments on an image retrieval experiment show that the average distance to the nearest image in the dataset can drop 40\% by using seGANs, which is a significant improvement (arguably more important than the usage of ensembles for discriminative networks). These results should be verified on multiple datasets. For future work we are especially interested to combine the ideas behind the cGANs with seGANs to further improve results. We are also interested to further improve the image retrieval system, which can be used as a tool to evaluate the quality of image generation methods. It would beneficial for research in GANs if a toolbox of experiments exists which allows us to quantitatively compare generative methods. 

\minisection{Acknowledgments}
This work is funded by the Projects TIN2013-41751-P of the Spanish Ministry of Science and the CHISTERA M2CR project PCIN-2015-226 and the CERCA Programme / Generalitat de Catalunya. We also thank Nvidia for the generous GPU donation. 

{\small
\bibliography{sample}}
\end{document}